\documentclass[fleqn,10pt]{wlscirep}
\usepackage[utf8]{inputenc}
\usepackage[T1]{fontenc}
\usepackage{array}
\usepackage{graphicx}
\usepackage{float}
\usepackage{multirow}       
\usepackage{tabularx}

 \title{A harmonised dataset for Earth system foundation models}
 
\author[1,2,3*]{Carlos Rodriguez-Pardo}
\author[1,2,3]{Massimo Tavoni}
\affil[1]{Politecnico di Milano, Department of Management, Economics and Industrial Engineering, Milan, Italy}
\affil[2]{RFF-CMCC European Institute on Economics and the Environment, Milan, Italy}
\affil[3]{CMCC Foundation - Euro-Mediterranean Center on Climate Change, Lecce, Italy}

\affil[*]{carlos.rodriguezpardo.jimenez@gmail.com}

\begin{abstract}
  Foundation models for Earth systems have so far been trained primarily on physical climate and weather data, with limited representation of the human systems that both drive and respond to environmental change. The lack of a unified global training resource that combines climate, land, ocean, cryosphere, infrastructure, hazards, and socioeconomic data on a common grid hinders progress toward truly multimodal Earth system foundation models. We present WorldTensor, a harmonised global dataset that aligns hundreds of environmental and socioeconomic variables to a standardised 0.25$^\circ$ spatial grid and annual temporal framework. WorldTensor integrates reanalysis products, remote sensing, emissions inventories, land use reconstructions, hydrological observations, infrastructure and hazard datasets, and socioeconomic indicators within a single representation designed for machine learning workflows. To build the dataset, we regridded inputs across heterogeneous native resolutions and projections, rasterised point and vector datasets into spatially meaningful gridded fields, and reconciled temporal coverages ranging from daily observations to sparse multiyear socioeconomic snapshots. All outputs are distributed as NetCDF files with standardised coordinates, variable metadata, and a common CF metadata convention. WorldTensor provides a reproducible resource for training and evaluating foundation models that learn coupled dynamics across environmental and human systems at planetary scale.
  \end{abstract}

\begin{document}

\flushbottom
\maketitle

\thispagestyle{empty}

\section*{Background \& Summary}

Foundation models for the Earth system have advanced rapidly for weather~\cite{lang2024aifs} and climate~\cite{nguyen2023climax}, where dense reanalysis products and satellite observations provide gridded, physically consistent training data. These resources have enabled models that learn atmospheric~\cite{bodnar2025foundation} and oceanic~\cite{epicoco2025medformer} dynamics at global extent. Most existing pipelines, however, remain centred on the physical climate system alone, omitting the human systems that shape emissions, land conversion, infrastructure, exposure, and vulnerability. We use \textit{foundation model} in the standard sense of a large model pretrained on broad data and adaptable to many downstream tasks. In the Earth sciences this spans sub-daily weather and climate emulators~\cite{lang2024aifs,nguyen2023climax,bodnar2025foundation} and, increasingly, geospatial and Earth-system representation models that learn transferable spatial features across domains~\cite{agarwal2024general}. WorldTensor targets the latter class: rather than emulating fast atmospheric dynamics, it supports models that learn coupled human-environment structure at annual resolution: for example, general-purpose geospatial embeddings, transfer across environmental and socioeconomic prediction tasks, downscaling or gap-filling of sparse socioeconomic fields, and feature generation for climate-impact and risk models.

This is fundamentally a data problem. The information needed to describe coupled human-earth dynamics already exists across reanalyses, remote sensing, land use reconstructions, emissions inventories, infrastructure databases, hazard catalogues, and socioeconomic indicators; but these remain fragmented across incompatible grids, projections, temporal frequencies, formats, and metadata conventions. Some are regularly sampled rasters; others are irregular point or vector geometries. Some are available daily or monthly; others only at sparse multiyear intervals. Even simple cross-domain analyses require substantial preprocessing before variables can be compared or ingested into machine learning workflows.

This fragmentation constrains the next generation of foundation models. More generally, it limits the capacity to inform societally relevant decisions at the interface of environmental and human systems. Climate risk assessment, Earth system prediction, and global policy increasingly demand models that reason across environmental and human processes jointly. Physical climate fields alone cannot capture energy system structure, economic geography, or cumulative disaster exposure; socioeconomic data detached from climate and land cover provide only a partial view. A unified representation is needed for models to learn not only how the Earth evolves, but how human systems modify and respond to those dynamics. Early multimodal efforts exist~\cite{agarwal2024general}, but remain centred on the Global North, limiting their broader utility.

WorldTensor aims to address these gaps, responding to recent calls for unified AI frameworks that integrate physical climate modelling, impact assessment, and socioeconomic analysis~\cite{ou2026artificial}. It harmonises hundreds of global environmental and human-system variables onto a shared $0.25^\circ$ latitude--longitude grid with standardised coordinates, consistent metadata, and a NetCDF representation designed for machine learning. Rather than organising data by source convention, WorldTensor treats the final product as a multimodal tensor spanning climate, extremes, emissions, air quality, land use, vegetation, hydrology, cryosphere, ocean biogeochemistry, agriculture, energy infrastructure, human systems, hazards and conflict, and static geographic covariates.

Building such a dataset requires more than format conversion. Sources differ in native resolution, projection, temporal structure, and geometry type. WorldTensor therefore regrids raster products from heterogeneous source grids, rasterises point and vector datasets into continuous fields, aggregates subannual observations to a common annual unit, and standardises coordinates, units, and metadata across all variables. Particular attention was paid to geospatial problems that silently degrade global products: longitude seam handling, polar edge behaviour, and the conversion of event or infrastructure datasets into gridded representations that preserve spatial signal without introducing artefacts.
The resulting dataset is intended as research infrastructure rather than a benchmark tied to one model class. It supports multimodal pretraining, transfer learning across Earth and human-system domains, coupled climate--society analysis, and feature generation for impact modelling. All outputs share a consistent file structure with provenance metadata and are accompanied by PyTorch ingestion scripts. More broadly, WorldTensor builds upon the view that planetary change emerges from the interaction of natural and human systems: emissions alter atmospheric composition, land use conversion modifies hydrology, infrastructure mediates exposure and adaptation, and hazard impacts depend jointly on environmental forcing and social vulnerability. Co-locating these dimensions in a single resource gives models the raw material to learn from these coupled dynamics.

\section*{Methods}

\subsection*{Dataset design principles}
WorldTensor is built around four principles: shared spatial support, shared temporal resolution, multimodal domain coverage, and a machine-readable format. Interoperability is prioritised over preserving each source in its native schema.

The spatial support is a regular $0.25^\circ$ global latitude--longitude grid, matching the ERA5 reanalysis~\cite{hersbach2020era5}. This balances geographic detail, global completeness, and computational tractability: a finer grid would increase storage and sparsity for variables derived from point or event data, while a coarser grid would blur important spatial gradients.

All non-static variables are represented at annual resolution. Daily and monthly products are summarised using variable-appropriate statistics; socioeconomic datasets at sparse intervals are interpolated from anchor years. This shared temporal unit preserves long-term evolution, enables cross-domain alignment, and avoids embedding source-specific irregularities. Time-invariant variables such as bathymetry are released as static layers. All outputs are packaged as compressed NetCDF files with standardised coordinates, variable-level metadata, and stable naming conventions. NetCDF is self-describing, efficient for large gridded arrays, and widely supported in Earth science and machine learning stacks such as \texttt{xarray}.

\subsection*{Source dataset selection}
Source selection followed scientific, practical, and operational criteria. The primary requirement was relevance to coupled Earth--human system analysis: each candidate was evaluated on whether it described a major component of planetary change: physical climate, land and ocean processes, emissions, infrastructure, hazards, or socioeconomic activity. We assemble a coherent set of high-quality variables that capture environmental states, human pressures, and human responses, rather than compile an exhaustive catalogue. A second criterion was global coverage. Sources with only regional extent or highly fragmented release structures were excluded. Temporal usability was equally important: included sources needed to be compatible with annual outputs, either natively or through aggregation. Long time series were preferred, but coverage differs across domains. Sources with poorly documented timestamps or no reliable temporal alignment method were excluded. Reproducibility was a further requirement. Preference was given to sources accessible through public APIs, stable endpoints, or versioned releases. The repository supports automated retrieval, scripted downloads, and manual staging for datasets subject to authentication or redistribution constraints. The processing framework separates raw acquisition from harmonised outputs and records provenance explicitly, so that reuse can follow the terms of each underlying dataset. Table~\ref{tab:sources} summarises the source collections, reporting released rather than full native coverage.

The resulting list is organised around the causal structure of coupled human--Earth dynamics rather than around data availability: it spans environmental states (climate, land, ocean, cryosphere, and vegetation), the human pressures that modify them (emissions, land-use change, and energy and settlement infrastructure), and the exposure and response variables that close the loop (socioeconomic indicators, hazards, and conflict). Each domain is included because it contributes a distinct axis of this system, and we deliberately favour breadth across these axes over exhaustive depth within any single one. Assembling them in a common representation is what allows models to learn cross-domain couplings (for example between emissions and atmospheric composition, or between land-use change and hydrology) that domain-specific datasets cannot express. The cost of this breadth is heterogeneity: the included sources differ in native resolution, temporal coverage, and measurement reliability (Tables~\ref{tab:sources}, \ref{tab:regrid}, and \ref{tab:quality}), some domains are considerably sparser than others, and co-locating many variables on a single grid can induce spurious cross-domain correlations if used without care. We therefore intend WorldTensor as a curated but non-exhaustive dataset from which users select the variables appropriate to a given task, rather than as a corpus that must be ingested in full.

\begin{table*}[ht]
\centering
\footnotesize
\setlength{\tabcolsep}{4pt}
\renewcommand{\arraystretch}{1.15}
\begin{tabularx}{\linewidth}{>{\raggedright\arraybackslash}p{1.55cm} >{\raggedright\arraybackslash}X >{\raggedright\arraybackslash}X >{\raggedright\arraybackslash}X >{\raggedright\arraybackslash}p{1.55cm}}
\hline
Release component & External source datasets & Native support & Released WorldTensor content & Released coverage \\
\hline
Climate & ERA5 monthly reanalysis~\cite{hersbach2020era5} & Global monthly reanalysis on the target $0.25^\circ$ grid & 276 annual climate layers with mean or sum as the primary statistic plus standard deviation, minimum, and maximum & 1940--2025 \\
Extremes & SPI/SPEI drought fields~\cite{vicenteserrano2010spei}, HadEX3 land heatwaves~\cite{dunn2020hadex3}, and NOAA marine heatwaves~\cite{hobday2016mhw} & Monthly drought grids, annual land-extreme indices, and monthly SST-anomaly fields & 8 drought layers, 4 land-heatwave layers, and 2 marine-heatwave layers & 1901--2025 \\
Air quality & CAMS EAC4 atmospheric-composition reanalysis~\cite{inness2019cams} & Monthly atmospheric-composition reanalysis & 40 annual concentration and total-column statistic layers & 2003--2024 \\
Emissions & EDGAR v8.0 non-CO$_2$ and biogenic CO$_2$~\cite{crippa2024edgar}, CEDS shipping NO$_x$~\cite{hoesly2018ceds}, and ODIAC fossil CO$_2$~\cite{oda2018odiac} & Annual sectoral flux rasters and monthly emissions grids & 67 annual emissions layers spanning CH$_4$, N$_2$O, biogenic CO$_2$ by sector, shipping NO$_x$, and ODIAC fossil CO$_2$. EDGAR fossil CO$_2$ families are excluded because their IEA-derived data carry a CC~BY-NC-ND~4.0 license incompatible with open redistribution & 1970--2024 \\
Land use & LUH3 states and transitions, with LUH methodological lineage documented by the LUH2 description~\cite{hurtt2020luh2} & Annual global land-use fractions and transition fluxes & 14 state layers and 98 transition layers & 1900--2024 \\
Vegetation & MOD13C2 greenness~\cite{huete2002modis_vi}, MCD64A1 burned area~\cite{giglio2018mcd64a1}, and VODCA~\cite{moesinger2020vodca} & Monthly vegetation indices, tiled burned-area observations, and 10-daily vegetation optical depth & 10 annual vegetation layers & 2000--2025 \\
Agriculture & GGCP10 crop production~\cite{qin2024ggcp10}, crop-specific fertilizer maps~\cite{coello2025fertilizer}, and AGLW livestock density~\cite{du2025livestock} & Yearly crop rasters, annual nutrient-input rasters, and annual livestock rasters & 4 crop-production, 3 fertilizer, and 5 livestock layers & 1961--2021 \\
Hydrology & GRACE/GRACE-FO~\cite{tapley2004grace,landerer2020gracefo}, GLDAS~\cite{rodell2004gldas}, and WAD2M~\cite{zhang2021wad2m} & Monthly land-surface fields, and annual wetland-dynamics grids & 8 annual hydrology layers & 2000--2025 \\
Cryosphere & ESA Snow CCI~\cite{luojus2021globsnow}, ESA Permafrost CCI~\cite{obu2019permafrost}, and WGMS glacier fields~\cite{dussaillant2025glaciers} & Daily snow products, annual permafrost grids, and annual glacier fields & 13 annual cryosphere layers & 1976--2025 \\
Ocean & MODIS-Aqua chlorophyll-\textit{a}~\cite{nasa2022modisa_chl} & Monthly ocean-color product & 4 annual chlorophyll layers & 2010--2023 \\
Human systems & GPWv4 and UN WPP~\cite{ciesin2018gpwv4,un2024wpp}, Kummu GDP family~\cite{kummu2025gdppc}, Wang \& Sun GDP~\cite{wang2022gdp}, SectGDP30~\cite{shoji2025sectgdp}, inequality rasters~\cite{chrisendo2025inequality}, HDI~\cite{kummu2018hdi}, HNTL~\cite{li2020hntl}, urban extents~\cite{zhao2022urban_extents}, GHSL~\cite{pesaresi2024ghsl}, GISA~\cite{huang2021gisa}, WSF~\cite{marconcini2020wsf}, Human Footprint~\cite{mu2022hfp}, and HMv2024~\cite{theobald2025hmv2024} & Anchor-year rasters, multiband GeoTIFFs, epoch products, and annual rasters & 22 annual socioeconomic, settlement, and human-modification layers & 1972--2024 \\
Energy & Global Integrated Power Tracker~\cite{gem2026gipt} & Plant-level point records with commissioning and retirement dates & 52 annual power-infrastructure layers & 1900--2025 \\
Hazards \& conflict & UCDP GED~\cite{sundberg2013ucdp}, USGS ComCat~\cite{usgs2017comcat}, IBTrACS~\cite{knapp2010ibtracs}, NOAA significant volcanic eruptions~\cite{noaa2014volcanoes}, and GDIS~\cite{rosvold2021gdis} & Event and track catalogs & 28 annual event, distance, and cumulative hazard layers & 1900--2025 \\
Static context & GMTED2010~\cite{danielson2011gmted}, ETOPO 2022~\cite{noaa2022etopo}, SoilGrids~\cite{poggio2021soilgrids}, global soil texture~\cite{reynolds2000soiltex}, FLDAS vegetation classes~\cite{mcnally2017fldas}, distance to coast, rivers, and cities~\cite{nasa2009dist2coast,lehner2013hydrorivers,weiss2018accessibility} & Static rasters, categorical masks, and proximity surfaces & 99 static layers & static \\
\hline
\end{tabularx}
\caption{Major external source collections in WorldTensor. Coverage is reported for the released files, not the full native source span. In several workflows the released span is shorter because incomplete years were excluded or the release was capped at 1900.}
\label{tab:sources}
\end{table*}

\subsection*{Canonical data model}
All variables share a regular geographic grid with latitude spanning $-90^\circ$ to $90^\circ$ and longitude spanning $0^\circ$ to $359.75^\circ$ at $0.25^\circ$ increments, using standardised coordinate names (\texttt{lat}, \texttt{lon}, and an annual \texttt{time} coordinate for temporal variables). This ensures all outputs can be combined without reprojection.

Temporal variables are stored as one file per variable per year under \path{<domain>/<variable>/<YYYY>.nc}. The one exception is \texttt{land\_use}, split into \path{states/} and \path{transitions/} subtrees. Static variables are stored under \path{static/<group>/<variable>.nc}. This modular layout simplifies partial reprocessing and selective access.

Each file is self-describing: data variables carry \texttt{units} and \texttt{long\_name} attributes, and global metadata include a CF conventions tag and source provenance. Outputs are compressed \texttt{float32}. A configuration registry (\texttt{config/variables.yml}) maps each variable to its canonical path. Names combine a short source code with a statistic suffix (e.g.\ \texttt{t2m\_mean}, \texttt{tp\_sum}). Where source naming could collide, domain-specific qualifiers are retained. The data model defines a consistent release abstraction rather than preserving every feature of each native schema.

\subsection*{Data acquisition and preprocessing}

Raw data are staged under source-specific directories via dataset-specific download modules and YAML configuration files. Acquisition uses scripted API requests, direct archive downloads, or manual file placement, depending on the source. This staged design separates acquisition from harmonised output generation and allows reprocessing without re-downloading.

Before harmonisation, each dataset undergoes source-specific preprocessing: variable selection, temporal parsing, coordinate normalisation, longitude standardisation, and missing-value identification. Gridded products may additionally require archive unpacking, band extraction, scale-factor application, and quality-flag propagation. High-frequency products are grouped into annual collections for consistent statistic derivation; sparse anchor-year datasets have their interpolation logic established before gridding. Point, line, and polygon datasets require geometry validation, coordinate standardisation, event-year parsing, and quantity selection before rasterisation.

\subsection*{Spatial harmonisation}

We convert all gridded inputs to the canonical WorldTensor grid while preserving each source's main scientific signal. Before resampling, coordinate names and ordering are standardised, latitude axes are oriented consistently, longitude conventions are normalised to a common $0$--$360^\circ$ system, and duplicated coordinates or invalid fill values are removed. For raster products carrying valid CRS metadata, reprojection is performed in a CRS-aware manner.  Global products require special care near the antimeridian and the poles. WorldTensor applies periodic handling of longitude to reduce seam artifacts and enforces a consistent south-to-north latitude ordering before interpolation. Continuous fields are generally harmonised using bilinear resampling, whereas discrete or categorical layers use nearest-neighbor or category-specific conversion logic. Table~\ref{tab:regrid} summarises the harmonisation method applied to each source. Continuous fields are regridded with bilinear interpolation; categorical layers use nearest-neighbour assignment to preserve class boundaries; and products substantially finer than the target grid are aggregated with area-weighted (conservative) averaging to avoid aliasing. Point and line datasets are not interpolated but rasterised by direct cell assignment or per-cell density.

\begin{table*}[ht]
\centering
\footnotesize
\setlength{\tabcolsep}{4pt}
\renewcommand{\arraystretch}{1.15}
\begin{tabularx}{\linewidth}{>{\raggedright\arraybackslash}p{4.2cm} >{\raggedright\arraybackslash}p{2.1cm} >{\raggedright\arraybackslash}p{1.9cm} >{\raggedright\arraybackslash}X}
\hline
Source / product & Native resolution & Field type & Harmonisation to the $0.25^\circ$ grid \\
\hline
ERA5 climate & $0.25^\circ$ & Continuous & Coordinate and longitude standardisation (native grid; no resampling) \\
CAMS EAC4 air quality & $0.75^\circ$ & Continuous & Linear interpolation with nearest-neighbour gap fill; periodic longitude \\
SPI/SPEI drought & $0.5^\circ$ & Continuous & Bilinear \\
HadEX3 land heatwaves & $\sim$1.25--1.9$^\circ$ & Continuous & Bilinear; linear interpolation across missing years \\
NOAA marine heatwaves & $1^\circ$ & Continuous & Bilinear; periodic longitude \\
EDGAR CH$_4$/N$_2$O/bio-CO$_2$ & $0.1^\circ$ & Continuous flux & Bilinear (fluxes clipped $\geq 0$) \\
CEDS shipping NO$_x$ & $0.5^\circ$ & Continuous flux & Bilinear \\
ODIAC fossil CO$_2$ & $1^\circ$ & Continuous flux & Bilinear \\
LUH3 states \& transitions & $0.25^\circ$ & Continuous fractions & Seam-padded bilinear; state fractions rescaled to local land budget \\
MOD13C2 NDVI/EVI & $0.05^\circ$ & Continuous & Area-weighted (conservative) block averaging \\
MCD64A1 burned area & 500\,m & Continuous (event) & Area-weighted averaging; summed to annual \\
VODCA vegetation optical depth & $0.25^\circ$ & Continuous & Bilinear (re-orientation/alignment) \\
GGCP10 crops, AGLW livestock, fertilizer & 5--10\,km & Continuous & Bilinear \\
GRACE(-FO), GLDAS, WAD2M & $0.25$--$0.5^\circ$ & Continuous & Bilinear \\
ESA Snow CCI & $0.1^\circ$ & Continuous & Bilinear; linear interpolation across missing years \\
ESA Permafrost CCI & 0.01$^\circ$ ($\sim$1\,km) & Continuous & Area-weighted averaging then bilinear \\
WGMS glaciers & $0.5^\circ$ & Continuous & Bilinear; periodic longitude \\
MODIS-Aqua chlorophyll-\textit{a} & 4\,km & Continuous & Linear interpolation; periodic longitude \\
GPW population, SectGDP30 & 2.5$'$ / 30$''$ & Continuous & Bilinear + inter-anchor temporal interpolation \\
Kummu/Wang GDP, GNI, HDI, inequality & $\sim$5$'$--0.5$^\circ$ & Continuous & Bilinear \\
Harmonized nighttime lights & $\sim$30$''$ & Continuous & Bilinear \\
GHSL, GISA, WSF, Human Footprint, HMv2024 & 30\,m--1\,km & Continuous / presence-year & Area-weighted averaging (GHSL, HMv2024, Human Footprint); first-detection-year products (WSF, GISA) converted to annual fractional coverage; masks nearest; epoch/anchor products (GHSL, HMv2024) linearly interpolated \\
GMTED2010 topography, ETOPO2022 bathymetry & 30$''$ / 60$''$ & Continuous & Area-weighted averaging then bilinear \\
SoilGrids properties & 250\,m & Continuous & gdalwarp coarsening (average) then bilinear \\
GLDAS soil texture, FLDAS vegetation class & native & Categorical & Nearest-neighbour (one-hot) \\
Distance-to-coast, travel-time-to-cities & 0.01$^\circ$ / 1\,km & Continuous & Bilinear (distance-to-coast subsampled from native, then bilinear) \\
HydroRIVERS & Vector (lines) & Line & Rasterised to per-cell presence; distance-to-river proximity surface \\
GIPT power plants & Point & Point & Direct cell assignment (capacity, counts) + spherical distance-to-nearest \\
Hazards \& conflict (UCDP, ComCat, IBTrACS, volcanoes, GDIS) & Point / track & Point & Direct cell assignment (counts, attribute sums) + spherical distance-to-nearest \\
\hline
\end{tabularx}
\caption{Spatial harmonisation method applied to each source. Continuous fields use bilinear interpolation; fields substantially finer than the target grid use area-weighted (conservative) averaging; categorical layers use nearest-neighbour assignment; and point and line datasets are rasterised rather than interpolated.}
\label{tab:regrid}
\end{table*}

\subsection*{Rasterisation of point, line, and polygon datasets}

Many input data sources in WorldTensor are published as points, lines, or polygons rather than rasters. To harmonise them into the shared format, we convert these geometries into annual grid-based representations using geometry-specific rules. Point datasets are typically assigned to grid cells and summarised as counts, sums, cumulative quantities, active stocks, or related metrics. Where simple cell aggregation is insufficient, additional continuous fields such as distance-to-nearest-event or accessibility-style surfaces are derived from the point distribution. Linear and polygonal datasets are handled using analogous but geometry-appropriate workflows. Line features are converted into per-cell length or density estimates by intersecting them with the canonical grid in a projected metric coordinate system. Polygonal sources are rasterised into masks, classes, coverage fields, or proximity surfaces according to the semantics of the original dataset. Across all geometry types, rasterisation is treated as a substantive representation step intended to produce spatially interpretable fields that can be analysed jointly with the gridded environmental variables.

\subsection*{Temporal harmonisation}

Temporal harmonisation converts heterogeneous source frequencies into a common annual release model. Daily and monthly inputs are aggregated to annual layers using variable-appropriate summary statistics, and workflows that require a full annual cycle exclude incomplete years rather than silently aggregating partial records. For datasets published only at sparse anchor years, intermediate annual layers are generated using explicit interpolation or year-assignment rules defined at the dataset level. Static variables are kept outside the annual framework and released as standalone layers without a time dimension. Processing years are always constrained by actual source availability, so missing or unpublished years are omitted rather than backfilled. In several workflows the released time span is intentionally shorter than the native source span because long historical sources are capped at the year 1900 for release consistency.

\subsection*{Domain-specific processing workflows}

\subsubsection*{Climate and climate extremes}

Climate fields are derived primarily from ERA5 monthly reanalysis~\cite{hersbach2020era5}. Because ERA5 is already published on the target $0.25^\circ$ latitude--longitude grid, this workflow mainly standardises coordinates and longitude convention rather than performing a full reprojection. For each variable and year, the primary annual statistic is computed together with the inter-monthly standard deviation, annual maximum, and annual minimum. Temperature and pressure variables use the annual mean as their primary statistic, while accumulative variables such as precipitation use the annual sum. This yields a consistent family of yearly climate layers of the form \texttt{\{variable\}\_\{stat\}} that retains multiple summaries of the seasonal structure. Climate extreme indicators are constructed from several sources. Drought indicators comprise SPI and SPEI at 1, 3, 6, and 12 month accumulation windows~\cite{vicenteserrano2010spei}. Monthly drought fields are normalised to the common coordinate convention, interpolated to the canonical grid, and reduced to annual mean surfaces. HadEX3 land heatwave indices are normalised to the common longitude convention, reindexed to a complete annual axis over the available period, and interpolated across missing years before final gridding~\cite{dunn2020hadex3}. Marine heatwaves are derived from NOAA monthly sea surface temperature anomaly fields and their corresponding monthly $90^\mathrm{th}$ percentile thresholds~\cite{hobday2016mhw}.

\subsubsection*{Emissions and air quality}

Anthropogenic emissions are harmonised from both annual and monthly inventories. EDGAR non-CO$_2$ greenhouse gases (CH$_4$, N$_2$O) and biogenic CO$_2$ are ingested as yearly substance/sector rasters at $0.1^\circ$ resolution~\cite{crippa2024edgar}. After converting longitudes to the $0$--$360^\circ$ convention, the rasters are regridded to the WorldTensor grid and clipped to non-negative fluxes. The workflow preserves the original sectoral decomposition, so each released layer corresponds to a specific gas/sector pair, and additionally supports aggregate sectors such as total aviation by summing the relevant components. EDGAR fossil CO$_2$ layers derived from IEA energy statistics are excluded from the release because they carry a \textit{CC~BY-NC-ND~4.0} license that prohibits derivative works. Where sources are published monthly, annual summaries are computed only from complete years. The CEDS shipping NO$_x$ product is regridded and then summarised into annual mean, standard deviation, minimum, and maximum layers~\cite{hoesly2018ceds}. ODIAC fossil CO$_2$ products are harmonised separately into annual total components: monthly fields are regridded to the target grid and then reduced to annual mean, sum, standard deviation, minimum, and maximum statistics~\cite{oda2018odiac}. Atmospheric composition variables are treated separately from emissions. CAMS global reanalysis monthly means are regridded, then converted to annual concentration or total-column summaries alongside dispersion and extremal statistics~\cite{inness2019cams}. This distinction allows emission fluxes and atmospheric state variables to coexist in the release without conflating source type or physical meaning.

\subsubsection*{Land use, vegetation, and food systems}

Land--use dynamics are represented through LUH3~\cite{hurtt2020luh2}. The states product is processed as annual fractional land-cover and land-management layers. Variables such as primary forest, secondary vegetation, pasture, rangeland, urban area, and crop functional types are extracted from the source NetCDF, padded across the longitude seam,  interpolated to the canonical grid, and written year by year. For grouped fractional states, the workflow rescales interpolated fractions to maintain a coherent local land budget. LUH3 transitions are handled separately as annual flux variables describing conversion between land--use states.

Vegetation products combine satellite greenness, burned area, and vegetation structure. MOD13C2 monthly NDVI and EVI are clipped to valid ranges, downscaled from $0.05^\circ$ to $0.25^\circ$ through area-weighted block averaging, and then aggregated to annual mean, standard deviation, maximum, and minimum fields~\cite{huete2002modis_vi}. MODIS burned area products are mosaicked from tiles, warped to the canonical grid, converted to approximate burned area per cell, and annualised as yearly sums to preserve event accumulation rather than average state~\cite{giglio2018mcd64a1}. VODCA L-band vegetation optical depth is already near the target resolution but differs in orientation and alignment. It is harmonised to annual mean layers after temporal aggregation of the 10-daily sources~\cite{moesinger2020vodca}. Together these products provide complementary views of vegetation productivity, disturbance, and canopy structure.

Agricultural layers are generated from multiple source types. GGCP10 crop production data are processed as yearly high resolution GeoTIFFs and regridded to annual production totals for major staple crops~\cite{qin2024ggcp10}. AGLW livestock rasters are processed analogously but preserve density units\cite{du2025livestock}. Fertilizer variables are constructed from annual 5-arcminute application rate rasters: all crop layers for a given nutrient and year are summed, and the resulting nitrogen, phosphorus, and potassium fields are regridded to the common grid~\cite{coello2025fertilizer}. These layers complement the crop production and livestock families with explicit management signals relevant to food systems.

\subsubsection*{Hydrology, cryosphere, and ocean}

Hydrological datasets combine terrestrial water storage, soil moisture, inundation, and snow products. GRACE and GRACE-FO data are aggregated from monthly liquid water equivalent thickness anomalies to annual mean, maximum, minimum, and standard deviation layers~\cite{tapley2004grace,landerer2020gracefo}. GLDAS monthly land surface fields are aggregated to annual mean root-zone soil moisture and snow water equivalent and then aligned to the canonical grid~\cite{rodell2004gldas}. Wetland inundation from WAD2M is annualised into yearly mean and yearly maximum inundation fraction, reflecting the importance of both average wetness and peak seasonal extent~\cite{zhang2021wad2m}. Cryospheric processing distinguishes between snow, glaciers, and permafrost. ESA Snow CCI daily products are aggregated to annual mean, standard deviation, minimum, and maximum fields for snow water equivalent~\cite{luojus2021globsnow}. Glacier variables from WGMS are processed as annual mass change and area fields on the common grid~\cite{dussaillant2025glaciers}. ESA CCI permafrost layers are annual products over the northern high latitudes, warped to the global $0.25^\circ$ grid using area-based resampling and clipped to physically meaningful ranges before writing annual permafrost extent fraction and active-layer-thickness variables~\cite{obu2019permafrost}. Ocean biogeochemistry is represented by chlorophyll \textit{a} from MODIS-Aqua~\cite{nasa2022modisa_chl}. Monthly fields are standardised to the WorldTensor longitude convention, padded across the antimeridian, regridded to the canonical grid, and aggregated to annual mean, standard deviation, maximum, and minimum layers. This yields a compact but globally consistent marine productivity signal directly comparable with the terrestrial and atmospheric domains.

\subsubsection*{Human systems, energy, hazards, and conflict}

Human system variables are generated from both raster and non-raster inputs. Population layers from GPW are available only for anchor years, so each anchor raster is regridded to the canonical grid and then interpolated between anchor years to produce annual population count and population density surfaces~\cite{ciesin2018gpwv4,un2024wpp}  (\url{https://population.un.org/wpp/}). Land area is retained as a static contextual layer. Sectoral GDP from SectGDP30 is treated similarly using its published anchor years~\cite{shoji2025sectgdp}. Additional gridded socioeconomic products, including total GDP, GDP per capita, GNI per capita, HDI, and inequality indicators, are harmonised from multiband GeoTIFFs, yearly raster archives, and NetCDF time cubes~\cite{kummu2025gdppc,chrisendo2025inequality,kummu2018hdi}.

Harmonised nighttime lights are processed as yearly rasters spanning the DMSP and VIIRS eras~\cite{li2020hntl}. Several settlement products require reconstruction from non-annual formats: WSF Evolution and GISA encode the first year of urban or impervious presence, which is converted into annual fractional coverage time series after tile mosaicking. GHSL built-surface and built-volume products are warped from published epoch layers and interpolated between anchor years to create annual built-environment indicators~\cite{marconcini2020wsf,huang2021gisa,pesaresi2024ghsl,zhao2022urban_extents}. Human modification indicators include HMv2024 transport and accessibility components, interpolated between five-year anchors~\cite{theobald2025hmv2024}, and annual Human Footprint rasters aggregated from their native grids~\cite{mu2022hfp}. A further derived layer, ecosystem service value, is calculated by combining annual LUH3 land--use fractions and WAD2M inundation with biome-specific valuation coefficients~\cite{costanza2014esv}.

Energy layers are processed through rasterisation from point observations. The power plant workflow reads plant-level records from the Global Integrated Power Tracker~\cite{gem2026gipt} (\url{https://globalenergymonitor.org/projects/global-integrated-power-tracker/}) , standardises commissioning and retirement years, aggregates capacity directly to $0.25^\circ$ cells, and produces yearly fields for active, added, retired, cumulative retired, and net generating capacity. The same workflow derives secondary fields such as distance to the nearest active plant, clean-power accessibility, active plant counts, average plant size, capacity-weighted mean age, type-diversity entropy, and renewable-proximity advantage.

Hazard and conflict products are derived from yearly event catalogs. Earthquakes, tropical cyclones, volcanic events, general disaster inventories, and conflict events are normalised into tabular annual point datasets with standardised coordinates and event years~\cite{usgs2017comcat,knapp2010ibtracs,noaa2014volcanoes,rosvold2021gdis,sundberg2013ucdp}. They are then rasterised by direct assignment to grid cells and supplemented with spherical distance to nearest event fields. A land mask is applied where appropriate. The released layers comprise yearly and cumulative counts together with source-specific attribute sums: earthquake magnitude and depth, cyclone wind speed and pressure, volcanic explosivity and fatalities, disaster damage, and conflict fatalities. This approach preserves both localised event occurrence and broader spatial exposure gradients, allowing event-driven disturbances to be represented in the same gridded format as environmental state variables.

\subsubsection*{Static geographic and land-surface context}

Static contextual layers are processed separately from the annual time series but follow the same spatial and metadata conventions. Topographic variables such as mean elevation, elevation variability, and slope are resampled from GMTED2010~\cite{danielson2011gmted}, while ETOPO 2022 provides bathymetric elevation and ocean depth~\cite{noaa2022etopo}. Geography layers include signed distance to coast  (\url{https://oceancolor.gsfc.nasa.gov/resources/docs/distfromcoast/}) and distance to river, with river distance derived from HydroRIVERS~\cite{lehner2013hydrorivers}. Land area and travel time to cities are released as static surfaces because they function as fixed covariates rather than annual time series~\cite{weiss2018accessibility}.

Additional static land surface descriptors are derived from several sources. SoilGrids properties are downloaded for multiple depths and regridded to static profiles of soil chemistry and texture~\cite{poggio2021soilgrids}. The GLDAS soil-texture classification and FLDAS vegetation classes are converted to one-hot layers using nearest-neighbor interpolation, so that categorical land surface information can be represented in the same raster format as continuous variables without blurring class boundaries~\cite{reynolds2000soiltex,mcnally2017fldas}. These layers are static in the release data model, but they condition or contextualise many of the time-varying processes represented elsewhere in WorldTensor.

\subsection*{Metadata standardisation and file packaging}

All release products are written as NetCDF files with a consistent set of structural and descriptive conventions. Each data variable carries at least \texttt{units} and \texttt{long\_name} attributes, and dataset-level metadata include a CF conventions tag together with source and title information. Outputs are encoded as compressed \texttt{float32} arrays to balance storage and precision. Many pipelines also attach the nominal year and a source identifier to the global metadata.

The packaging strategy is modular. Temporal variables are written as one file per year inside a variable-specific directory, so the directory name serves as the stable machine-readable identifier and the file name provides the temporal key. Static variables are stored as standalone layers without an artificial time axis. Shared helpers such as \texttt{output\_path\_for}, \texttt{save\_annual\_variable}, and \texttt{save\_static\_variable} enforce the canonical layout. This organisation allows individual variable families to be updated, replaced, or inspected independently without rebuilding the full corpus.

\subsection*{Automated quality control}

Quality control is implemented both during writing and as a release-level audit. At write time, helper functions enforce a minimum metadata contract, standardised coordinate names, compressed \texttt{float32} storage, and the expected output layout. Several workflows also include sanity checks such as filtering invalid coordinates, rejecting incomplete years, clipping variables to physically meaningful ranges, and skipping files whose content is inconsistent with the requested processing interval.

After generation, the full corpus is subjected to automated structural auditing across all NetCDF files. The audit checks grid dimensions, coordinate names and values, longitude convention, time-dimension presence, data type, CF-style variable attributes, global metadata, compression, temporal bounds, and year continuity within each variable directory. An accompanying harmonisation script can then apply fixes such as adding missing time dimensions, patching absent CF attributes, casting variables to \texttt{float32}, regridding residual nonconforming files, and removing files outside the intended temporal range.

\subsection*{Quality flags and uncertainty}
Source products differ in the quality and uncertainty information they provide, and WorldTensor handles this at the preprocessing stage rather than as companion variables. Where sources carry per-pixel quality flags, these flags are used to mask low-quality or invalid pixels before aggregation (MODIS QA masking defaults to a moderate threshold), but the flags themselves are not retained in the release. Source-provided uncertainty layers, such as the SoilGrids quantile surfaces and the GRACE measurement-error and scale-factor fields, are not propagated into WorldTensor; the harmonised products retain the central estimate only. Throughout, variables are clipped to physically meaningful ranges, and every released file carries a finite-value mask that identifies missing or masked cells. Importantly, the annual standard-deviation, minimum, and maximum layers describe within-year temporal variability of the aggregated statistic, not measurement uncertainty. Table~\ref{tab:quality} summarises, per source class, the native quality and uncertainty information and its treatment. Users requiring formal per-pixel uncertainty should consult the native products listed in Table~\ref{tab:sources}.

\begin{table*}[ht]
\centering
\footnotesize
\setlength{\tabcolsep}{4pt}
\renewcommand{\arraystretch}{1.15}
\begin{tabularx}{\linewidth}{>{\raggedright\arraybackslash}p{4.4cm} >{\raggedright\arraybackslash}X >{\raggedright\arraybackslash}X}
\hline
Source class & Native quality / uncertainty information & Treatment in WorldTensor \\
\hline
ERA5, CAMS reanalyses & Model reanalysis; no per-pixel quality flags & Used as provided; no uncertainty propagated \\
MOD13C2 vegetation indices & Per-pixel QA / reliability bands & QA-bit masking applied before aggregation (moderate threshold); flags not retained \\
MCD64A1 burned area & Per-pixel QA and uncertainty bands & Only the valid burn-date range is used; QA and uncertainty bands dropped \\
ESA CCI Snow & Negative fill / flag values (non-valid categories) & Negative (invalid) values masked before aggregation; not retained \\
SoilGrids & Quantile (Q0.05/Q0.5/Q0.95) uncertainty layers & Only the mean is requested; quantile uncertainty not propagated \\
GRACE / GRACE-FO & Measurement-error and scaling-factor fields & Not propagated; annual mean/std/min/max retained \\
Emissions (EDGAR, CEDS, ODIAC) & No per-pixel uncertainty & No uncertainty propagated; EDGAR fluxes clipped to non-negative (ODIAC and CEDS retained as regridded) \\
LUH3 land use & Internal land-budget constraint & State fractions rescaled to the land budget; no per-pixel uncertainty \\
Gridded socioeconomic (GDP, HDI, inequality, settlement) & Generally none at pixel level & Physical-range clipping where a range applies (e.g.\ HDI, Gini, built-surface fraction) \\
Point catalogues (hazards, power plants) & Record-level attributes / metadata & Invalid coordinates and sub-threshold records dropped; records missing event/commissioning years excluded from the relevant counts; no per-event uncertainty retained \\
All temporal variables & --- & Released std/min/max capture within-year temporal variability, not measurement uncertainty; a finite-value mask flags missing cells \\
\hline
\end{tabularx}
\caption{Native quality and uncertainty information per source class, and its treatment in WorldTensor. Quality flags are applied as masks during preprocessing but not retained; source uncertainty layers are not propagated into the release.}
\label{tab:quality}
\end{table*}

\section*{Data Records}

\subsection*{Release structure and file organisation}
The WorldTensor dataset is deposited at Zenodo\cite{worldtensor_data} and is released as a modular collection of NetCDF files. The corpus contains 757 released units: 658 temporal variable families and 99 static layers, corresponding to 52{,}823 individual NetCDF files and approximately 46~GB on disk. For most temporal variables, the file layout follows \path{<domain>/<variable>/<YYYY>.nc}, with one annual layer per file and one variable family per directory. The one exception is \texttt{land\_use}, which is split into \texttt{states} and \texttt{transitions} subtrees. Static layers are stored as standalone files under \path{static/<group>/<variable>.nc}.

The relationship between released variable families and external source collections is summarised in Table~\ref{tab:sources}. The reported coverage refers to the files actually distributed in WorldTensor rather than the full native time span of every source product.

At a high level, the release can be read as a nested file schema in which the first directory denotes the broad domain, an optional intermediate level denotes a structured subtype, the next level denotes the canonical variable family, and the file name denotes the year:

\begin{verbatim}
  <domain>/
    <variable>/<YYYY>.nc

  land_use/
    states/<variable>/<YYYY>.nc
    transitions/<variable>/<YYYY>.nc

  static/
    <group>/<variable>.nc
\end{verbatim}

In this structure, \texttt{<domain>} corresponds to a scientific family such as \texttt{climate}, \texttt{emissions}, or \texttt{human\_systems}. The folders \texttt{states} and \texttt{transitions} are the only subtype folders in the release. The token \texttt{<variable>} is the canonical machine-readable identifier, and \texttt{<YYYY>.nc} stores a single annual layer for that variable family. This organisation makes WorldTensor easy to subset, update, and analyze without loading a monolithic archive.

\subsection*{Domain inventory}
WorldTensor is organised into 14 top-level domains: climate, extremes, air quality, emissions, land use, vegetation, hydrology, cryosphere, ocean, agriculture, energy, human systems, hazards and conflict, and static context. The largest domains by variable-family count are climate (276 families), land use (112 including states and transitions), static context (99 layers), and emissions (67). They are followed by energy (52), air quality (40), hazards and conflict (28), and human systems (22). Smaller but scientifically important domains include extremes (14), cryosphere (13), agriculture (12), vegetation (10), hydrology (8), and ocean (4).

The taxonomy is narrative rather than source-driven. Settlement, population, economic, and human modification products are consolidated under \texttt{human\_systems}. Atmospheric composition is separated from emissions under \texttt{air\_quality}. Natural hazards and conflict events are released together under \texttt{hazards\_and\_conflict}. This organisation reflects the scientific framing of WorldTensor as a coupled human--Earth system dataset.



\subsection*{Variable naming and metadata conventions}
Variable identifiers are derived from the parent directory names and serve as stable keys for discovery and downstream ingestion. For geophysical products, WorldTensor preserves a source short name plus an annual statistic suffix such as \texttt{mean}, \texttt{std}, \texttt{min}, \texttt{max}, or \texttt{sum} (e.g.\ \texttt{t2m\_mean} or \texttt{pm2p5\_std}). For more semantically specific domains, descriptive names are used, such as \texttt{population\_density}, \texttt{gdp\_total}, or \texttt{power\_active\_capacity\_mw\_total}. One deliberate exception is the EDGAR emissions inventory, where substance/sector identifiers (e.g.\ \texttt{ch4\_ags}, \texttt{n2o\_ene}) mirror the upstream sector taxonomy to preserve source provenance.

Each NetCDF file is self-describing. Data variables carry at least \texttt{units} and \texttt{long\_name} attributes, and files include standardised latitude and longitude coordinates, a \texttt{Conventions} attribute, and source metadata. Most temporal files represent a single annual layer using a singleton \texttt{time} coordinate, whereas static layers are stored without a time axis. This combination of stable directory-level identifiers and consistent file-level metadata allows the release to be discovered recursively and merged programmatically into larger tensors.

\subsection*{Temporal coverage and completeness}
WorldTensor spans 1900--2025 overall, but coverage differs substantially by domain and variable family because it is constrained by source availability. Climate variables span 1940--2025, land--use states 1900--2024, land--use transitions 1900--2023, energy 1900--2025, and hazards and conflict 1900--2025, although specific hazard families have shorter coverage such as conflict (1989--2024) and the GDIS disaster catalog (1960--2018). Human system variables span 1972--2024, hydrology 2000--2025, air quality 2003--2024, cryosphere 1976--2025, ocean 2010--2023, vegetation 2000--2025, and agriculture 1961--2021.

Coverage is also heterogeneous within domains. The \texttt{extremes} domain combines land heatwave indices from 1901--2018, marine heatwave metrics from 1991--2025, and SPI/SPEI drought indices from 1940--2025. The \texttt{emissions} domain combines EDGAR non-CO$_2$ and biogenic CO$_2$ series from 1970--2024, CEDS shipping NO$_x$ summaries from 1970--2017, and ODIAC fossil CO$_2$ products from 2000--2023. Within \texttt{human\_systems}, settlement layers range from 1972--2019 for impervious surface timing to 2000--2024 for annual Human Footprint. These differences are intentional and reflect source availability and completeness filtering, not an attempt to backfill a synthetic common date range.

\begin{figure}[!tb]
\centering
\includegraphics[width=\linewidth]{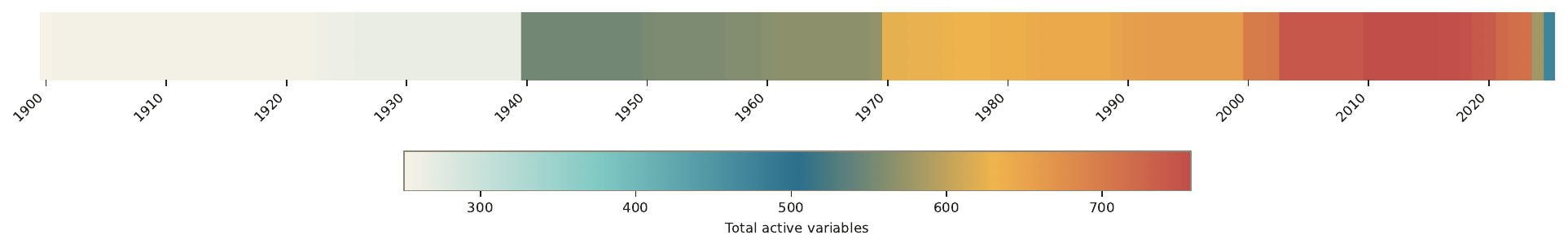}
\caption{Temporal density of WorldTensor, 1900--2025. Each column represents one year; colour encodes the total number of variable families with data available in that year, from roughly 270 in the early record to over 750 after 2000. The progressive saturation reflects the staggered onset of source datasets across domains.}
\label{fig:temporal_stripes}
\end{figure}

Figure~\ref{fig:temporal_stripes} shows that this heterogeneity is a property of the released data model rather than a packaging inconsistency.

\begin{figure}[!tb]
\centering
\includegraphics[width=\linewidth]{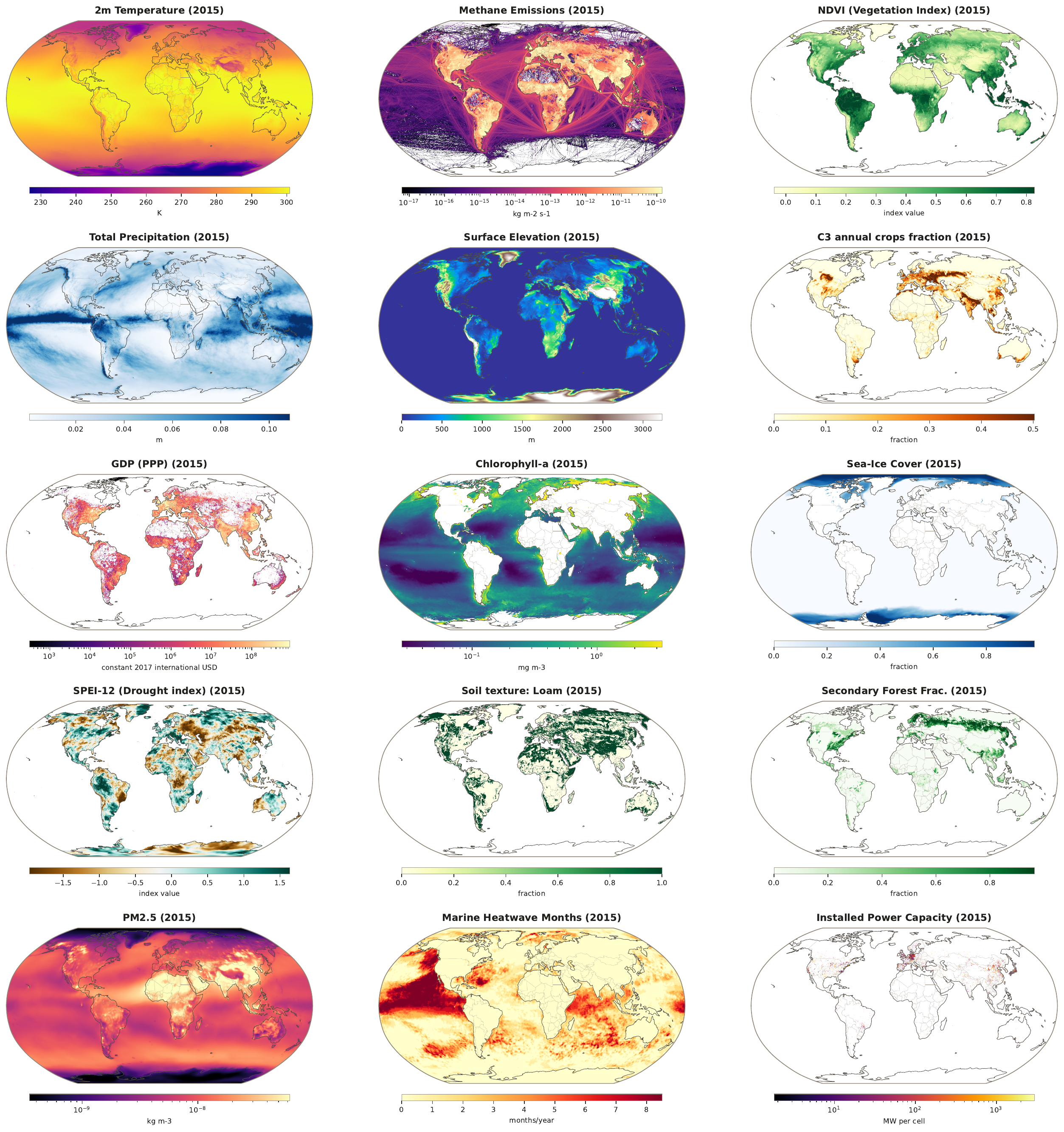}
\caption{Representative maps from WorldTensor for the year 2015. The gallery illustrates the spatial diversity of the dataset across physical climate fields, biogeochemical and land surface indicators, socioeconomic variables, infrastructure, and static contextual layers, all on the common 0.25$^\circ$ grid.}
\label{fig:spatial_gallery}
\end{figure}

Figure~\ref{fig:spatial_gallery} demonstrates that variables with very different scientific semantics can be compared directly on the shared grid.

\subsection*{Static layers}
The \texttt{static} domain contains 99 time-invariant layers grouped into eight subdomains: \texttt{bathymetry} (2), \texttt{geography} (2), \texttt{land\_area} (1), \texttt{soil\_properties} (60), \texttt{soil\_texture} (12), \texttt{topography} (3), \texttt{travel\_time\_to\_cities} (1), and \texttt{vegetation\_class} (18). These layers provide contextual information that either does not vary over time or is used as a fixed covariate across annual series. Examples include bathymetric elevation and ocean depth, distance to coast and distance to river, SoilGrids property/depth combinations, one-hot soil texture and vegetation class layers, and the static land area and travel time to cities surfaces. The current release emphasises physical and land surface context rather than ecological or administrative layers.

\subsection*{Structural conventions}
Two deliberate structural exceptions exist. First, \texttt{land\_use} retains one additional nesting level to separate \texttt{states} from \texttt{transitions}. Second, emissions families derived from EDGAR use lowercase sector codes (e.g.\ \texttt{ch4\_ags}, \texttt{n2o\_ene}) that mirror the upstream EDGAR sector taxonomy. A few domain assignments are narrative: Human Footprint, nighttime lights, settlement products, and economic layers are all grouped under \texttt{human\_systems} even though they originate from different source classes, and \texttt{travel\_time\_to\_cities} is packaged as a static layer. These choices reflect the intended conceptual organisation of WorldTensor.

\subsection*{Data access and recommended use}
The primary release artifact is the modular directory tree under \texttt{data/final}. Users can work directly with individual variable families by reading yearly NetCDF files from the relevant domain directories, or combine selected families into larger tensors by merging on the shared coordinates.

The code release includes lightweight PyTorch examples under \texttt{examples/torch/} that read directly from the per-variable layout. The script \texttt{01\_global\_tensor.py} demonstrates a \texttt{WorldTensorYearDataset} interface that stacks selected annual and static variables into aligned global tensors of shape \texttt{[C, H, W]}, together with coordinates, year labels, and finite-value masks. The companion script \texttt{02\_patch\_dataloader.py} demonstrates a \texttt{WorldTensorPatchDataset} interface that samples spatial crops from the same yearly stacks and can emit either dense patches or sparse dictionaries containing coordinates, values, masks, and grid indices. These examples provide a reference workflow for model pretraining, minibatch construction, and patch-based experimentation.

In downstream workflows, temporal series should be joined on the shared \texttt{time}, \texttt{lat}, and \texttt{lon} coordinates, while static layers can be broadcast across time as needed. The provided PyTorch examples handle this distinction automatically. Because temporal coverage differs by variable family, users should check coverage per variable or per domain before building training tensors.

\section*{Technical Validation}

We validated WorldTensor through a five-layer framework covering physical plausibility, internal consistency, temporal signal fidelity, spatial structure, and encoding compliance.

\subsection*{Physical plausibility and encoding}
Every variable was tested against physically motivated value bounds covering 54 representative variables across all 14 domains. All 54 passed with fewer than 0.1\% of pixels exceeding the expected range. The generated NetCDF files was checked for encoding compliance (float32 data type, latitude/longitude coordinate metadata): all files passed. Global area-weighted means were compared against five authoritative benchmarks: annual mean 2\,m temperature against the WMO State of Climate 2023 report, total precipitation against ERA5 documentation, NDVI against MODIS validation reports, GPWv4 population totals against the GPWv4 documentation, and anthropogenic CH\textsubscript{4} emissions against the EDGAR~v8.0 documentation. All five benchmarks fell within their tolerance bands.

\subsection*{Internal consistency}
Land-use state fractions from LUH3 (12~variables: primary and secondary forest, crops, pasture, rangeland, and urban) were evaluated against the expected land budget after regridding. Because the released active state layers exclude the static ice/water component, the relevant invariant is that their sum matches $1-\mathrm{icwtr}$ rather than unity. Across five test years (1950, 1980, 2000, 2010, 2020), land pixels show mean sums of 0.867 against mean expected budgets of 0.882, with mean absolute deviations of 0.015 and 92.7\% of land pixels within 0.02 of the target budget. Residual larger mismatches are localised to narrow coastal and island cells. Eleven cumulative hazard and infrastructure variables were verified to be monotonically non-decreasing over time, with zero violations detected.

\subsection*{Temporal signal fidelity}
\begin{figure}[H]
\centering
\includegraphics[width=.8\linewidth]{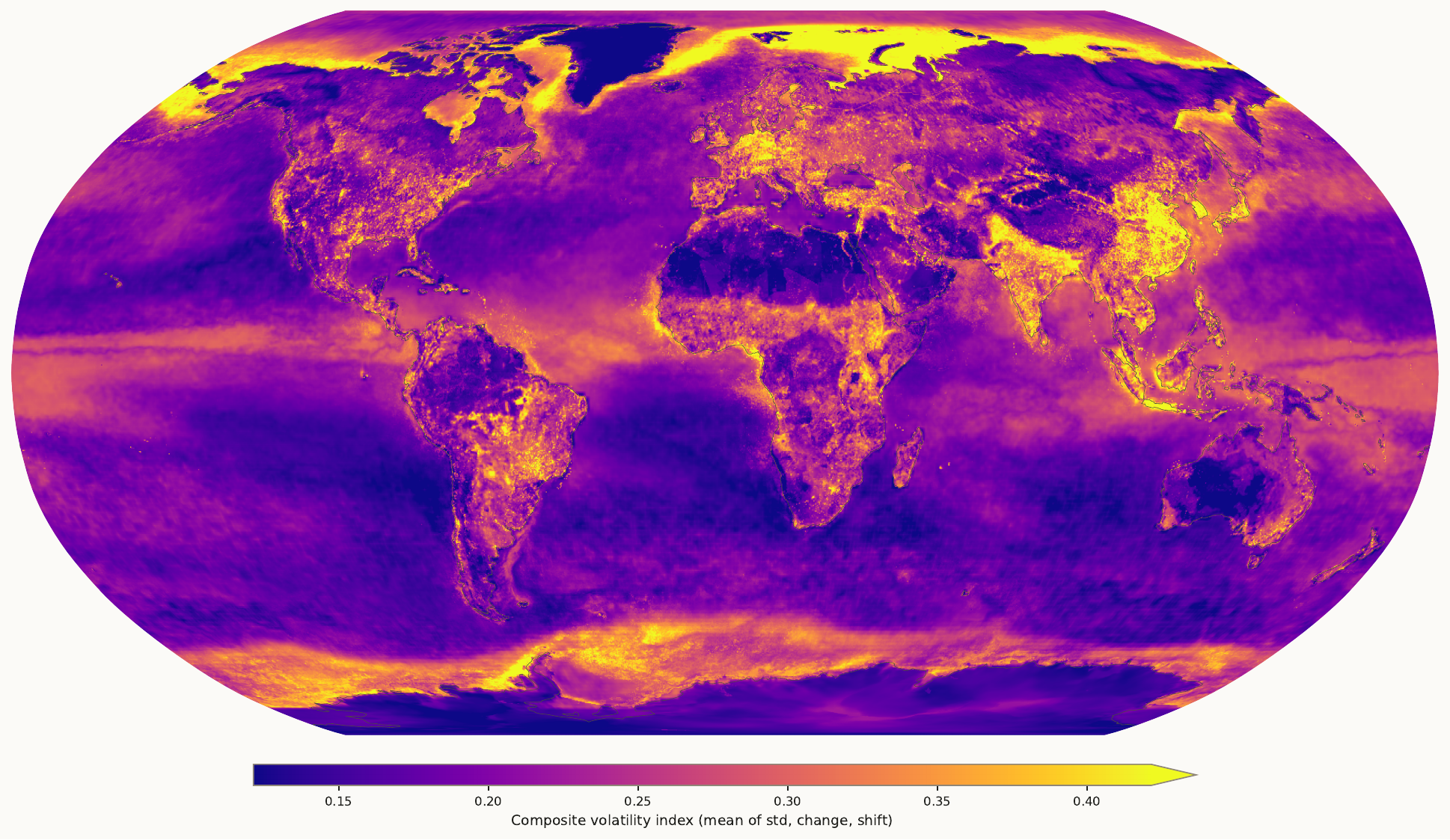}
\caption{Composite temporal volatility across all WorldTensor time-varying variables, computed as the cross-variable mean of normalised standard deviation, mean absolute year-to-year change, and terminal shift ($|\text{late mean} - \text{early mean}|$). Bright regions indicate grid cells where multiple variables changed substantially over the observational record. Hotspots include the Arctic (amplified warming, sea-ice loss), East and South Asia (urbanisation, land-use intensification), and equatorial ocean bands (ENSO variability).}
\label{fig:volatility}
\end{figure}

We tested whether WorldTensor captures five well-documented historical events without prior knowledge: the 1997--98 and 2015--16 El Ni\~{n}o episodes, the 1991 Pinatubo eruption, the 2020 COVID-19 emissions perturbation, and the 2012--2015 Syrian civil war. For each event, area-weighted regional or global mean time series were extracted and anomalies were detected via z-score analysis relative to the full series excluding the event window. Three of five events were detected at $|z| > 1$: the 2015--16 El Ni\~{n}o produced a clear temperature spike ($z = 2.1$), and the Syrian civil war produced extreme anomalies in both conflict event counts ($z = 6.1$) and fatalities ($z = 4.5$) over the target region. The Pinatubo eruption showed the expected direction in surface solar radiation ($z = -1.2$) but was attenuated in global annual mean temperature. The COVID-19 signal was similarly attenuated at annual resolution, consistent with the rapid within-year recovery reported by the Global Carbon Project. A composite temporal volatility map aggregating all time-varying variables (Figure~\ref{fig:volatility}) reveals that the Arctic, East and South Asia, and equatorial ocean bands exhibit the strongest multi-variable change over the observational record.

\subsection*{Spatial structure}
\begin{figure}[H]
\centering
\includegraphics[width=.8\linewidth]{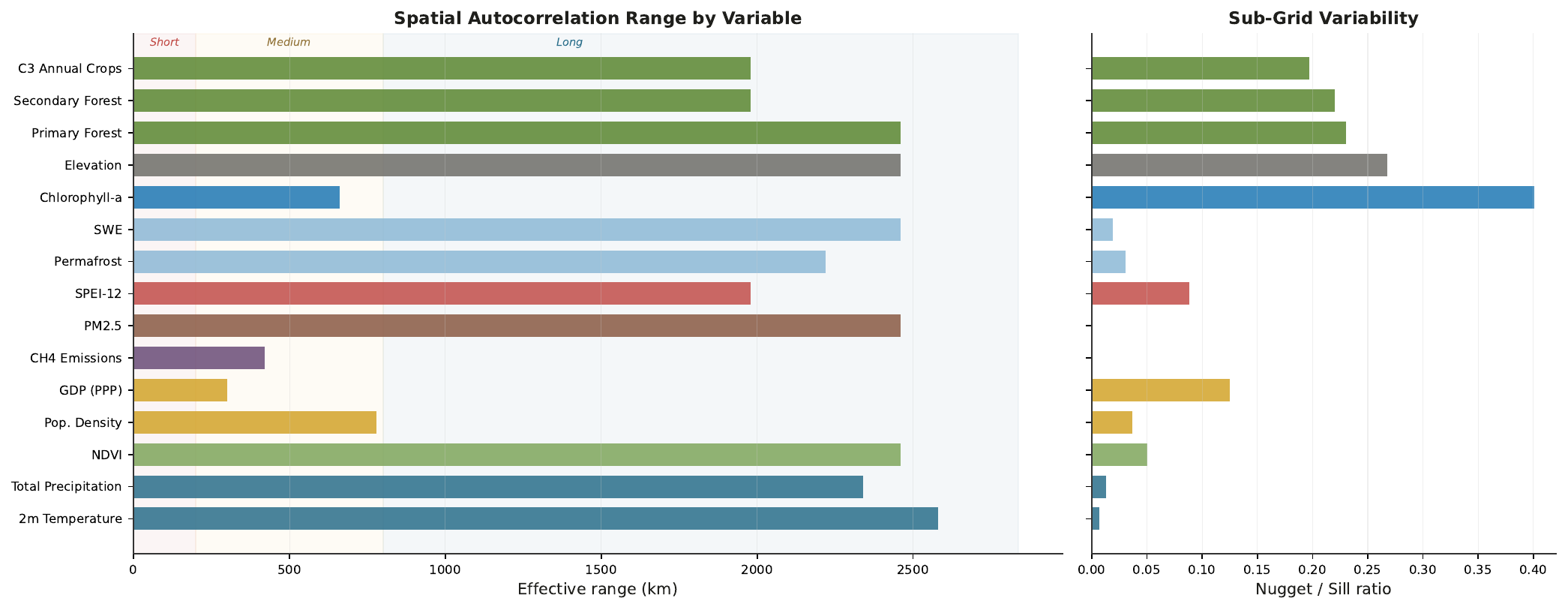}
\caption{Spatial autocorrelation summary for 15 representative WorldTensor variables. Left: effective range (km) of the fitted semivariogram, indicating the distance at which spatial correlation decays. Right: nugget-to-sill ratio, measuring the fraction of total variance attributable to sub-grid or measurement noise.}
\label{fig:variograms}
\end{figure}

Empirical semivariograms were computed for 15 representative variables spanning all major domains, confirming that broad spatial autocorrelation structure is preserved through regridding (Figure~\ref{fig:variograms}). Temperature exhibits long-range autocorrelation with an effective range of about 2580\,km. Precipitation and NDVI also retain broad spatial structure in this benchmark set, with fitted ranges of roughly 2340\,km and 2460\,km respectively. Among human-system variables, GDP remains distinctly localised (about 300\,km) and population density is shorter-range than most environmental fields (about 780\,km). Chlorophyll-\textit{a} shows the highest sub-grid variability (nugget/sill $\approx 0.40$), reflecting fine-scale oceanographic patchiness, whereas temperature has the lowest nugget-to-sill ratio ($<0.01$). PM\textsubscript{2.5} was the only benchmark variable flagged as atypical, with an anomalously long fitted range relative to its expected short-range class. Overall, the variogram fits support preservation of large-scale spatial structure across domains while highlighting a small number of variables that warrant more cautious interpretation.

\subsection*{Multivariate coherence via ICA}
\begin{figure}[H]
\centering
\includegraphics[width=.8\linewidth]{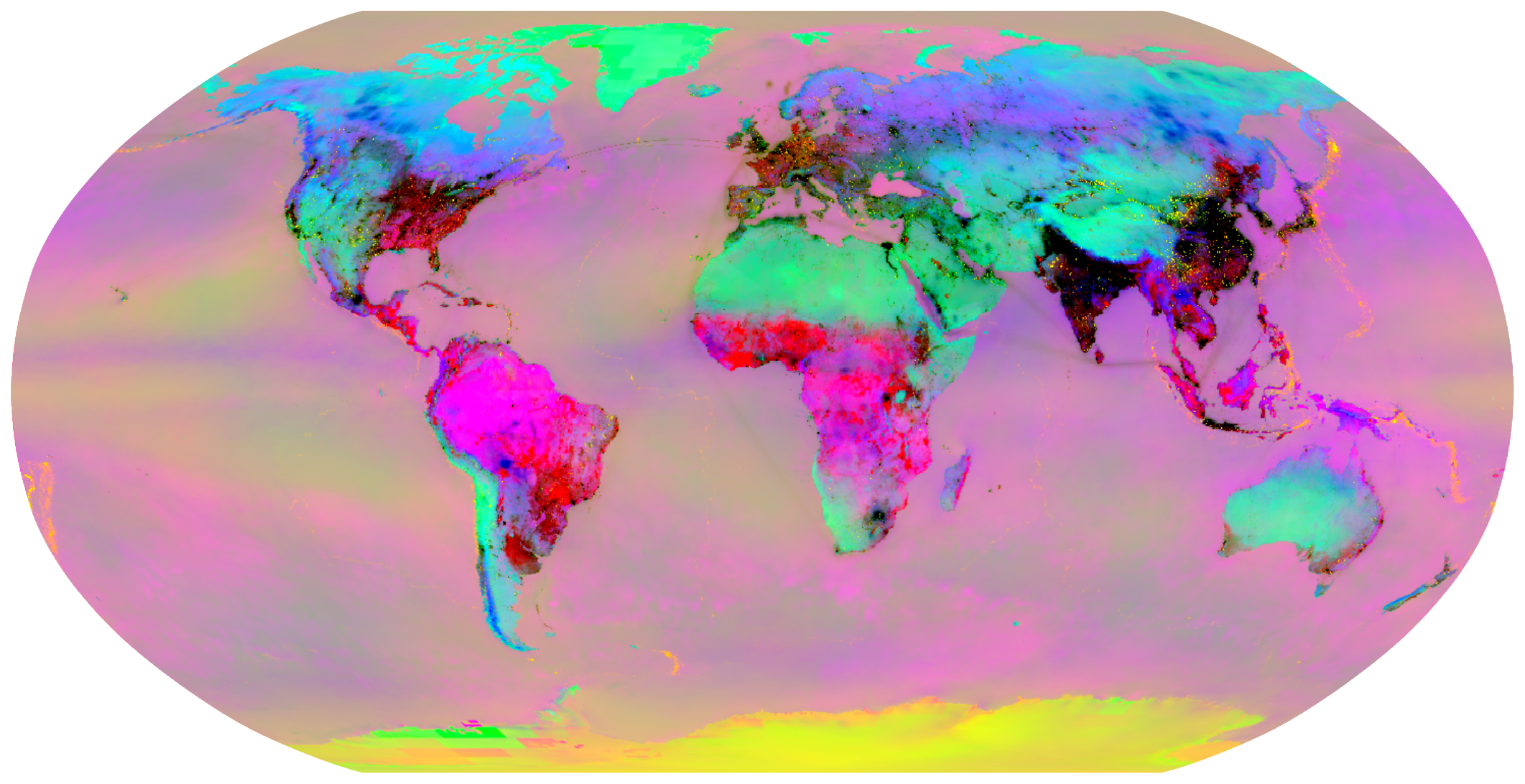}
\caption{Global composite of the first three ICA components derived from all non-static WorldTensor variables (2015), mapped to RGB. Distinct colours correspond to coherent eco-climatic regimes, confirming that cross-domain variables jointly capture geographically meaningful structure without supervision.}
\label{fig:ica_rgb}
\end{figure}

As a holistic check, we applied Independent Component Analysis (ICA) to a matrix of all non-static variables for the year 2015 and mapped the first three independent components to the red, green, and blue channels of a global composite (Figure~\ref{fig:ica_rgb}). Without any geographic supervision, the decomposition recovers recognisable eco-climatic zones: boreal forests, tropical rainforests, arid regions, and the ocean--land boundary emerge as distinct colour clusters. This confirms that WorldTensor's cross-domain variables jointly encode geographically meaningful structure and that no large-scale artefacts (e.g.\ tile seams or misaligned grids) contaminate the data.

\subsection*{Geospatial embedding analysis}
\begin{figure}[H]
\centering
\includegraphics[width=0.7\linewidth]{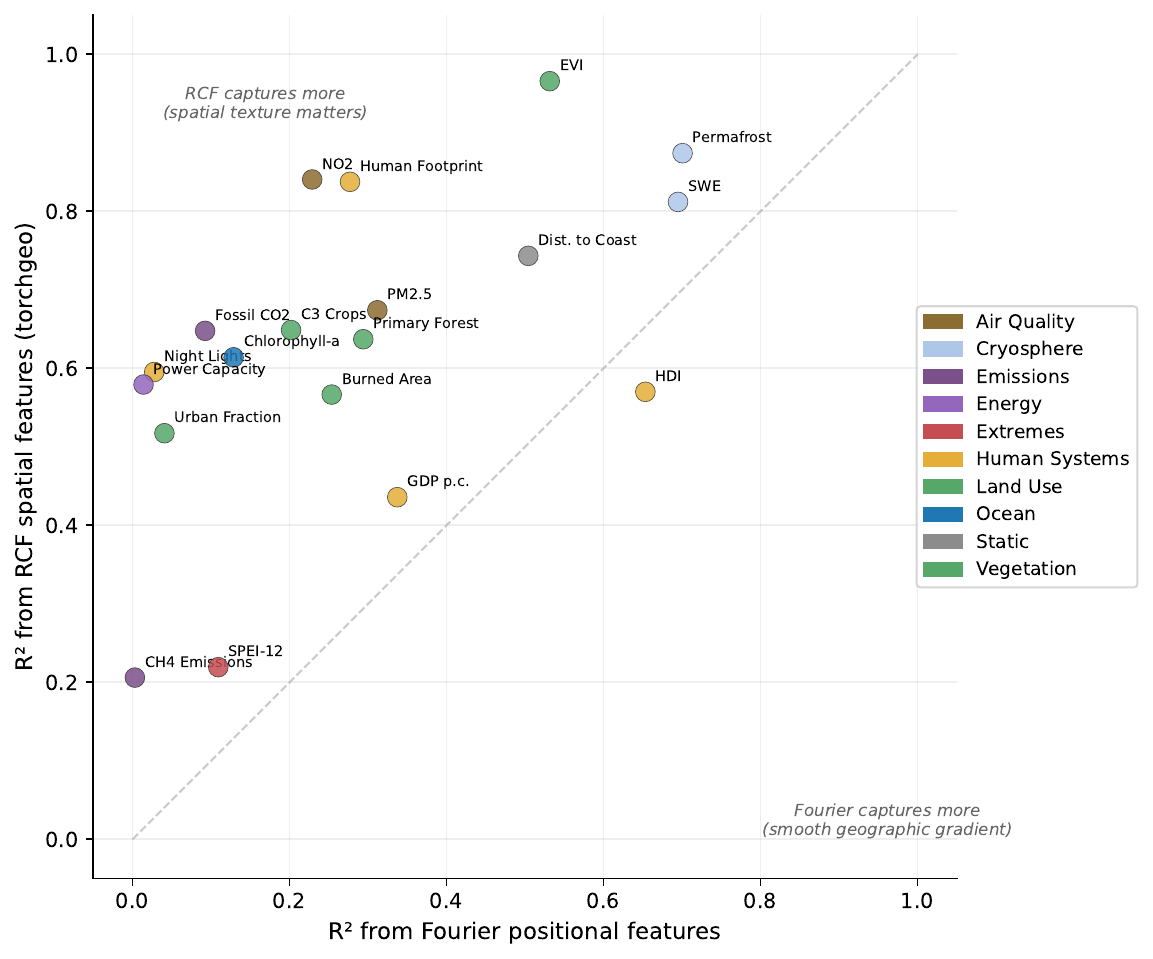}
\caption{Fourier positional encoding versus RCF spatial features (torchgeo) for the 19 WorldTensor targets eligible for RCF evaluation; the 7 variables used as RCF input channels are excluded to avoid information leakage. Points above the diagonal indicate variables where local spatial texture (captured by RCF) is more informative than geographic coordinates alone. Marker colour indicates domain.}
\label{fig:rcf_vs_fourier}
\end{figure}

To assess whether WorldTensor preserves meaningful spatial structure beyond simple geographic gradients, we compared three feature representations for predicting held-out variables via Ridge regression (Figure~\ref{fig:rcf_vs_fourier}). \textbf{Fourier positional encoding} (latitude/longitude transformed into 130 sin/cos features) captures smooth geographic gradients and achieves a mean test-set $R^2$ of 0.34 across 26 probe targets. \textbf{Cross-variable features} (14 climate and topographic predictors) improve this to $R^2 = 0.41$ across the same set, confirming physical coupling across domains. \textbf{Random Convolutional Features} (RCF), extracted from 7-channel spatial patches using the MOSAIKS approach~\cite{rolf2021mosaiks} implemented in torchgeo~\cite{stewart2022torchgeo}, are evaluated on the remaining 19 eligible targets after excluding those 7 input channels and achieve a mean $R^2$ of 0.63. The RCF features capture local spatial texture and multi-variable co-occurrence patterns that coordinates alone cannot encode. Variables with the largest RCF gain over Fourier features include EVI ($R^2_{\mathrm{RCF}} = 0.97$ vs.\ $R^2_{\mathrm{Fourier}} = 0.53$), NO\textsubscript{2} (0.84 vs.\ 0.23), and power plant capacity (0.58 vs.\ 0.01), demonstrating that WorldTensor preserves fine-grained spatial information across domains.

\section*{Usage Notes}

\subsection*{Machine learning workflows}
WorldTensor is designed for direct ingestion into machine learning pipelines. The code release provides two reference PyTorch interfaces under \texttt{examples/torch/}. The \texttt{WorldTensorYearDataset} builds full-resolution global tensors of shape \texttt{[C, H, W]} for a requested year and variable set, automatically distinguishing between temporal and static layers. Static variables are loaded once and broadcast across all requested years, while temporal variables are read from the corresponding annual file. Each sample includes the data tensor, a boolean finite-value mask, coordinate arrays, and the year label. The \texttt{WorldTensorPatchDataset} extends this interface to spatial cropping: it samples patches of configurable size from the global stacks and can return either dense arrays or sparse dictionaries containing coordinates, values, masks, and grid indices. Both interfaces handle missing data transparently through the finite-value mask, so downstream models can implement masking or imputation strategies as appropriate.

\subsection*{Static layers as geographic priors}
The 99 static layers in \texttt{static/} encode time-invariant properties of the Earth's surface: topography, bathymetry, soil composition, vegetation class, distance to coast and rivers, and travel time to cities. These layers can serve as geographic priors or grounding features in foundation model pretraining. A model pretrained on static context acquires a spatial understanding of the Earth's physical and land-surface composition before encountering temporal dynamics. This two-stage approach---first learning where mountains, coasts, soil types, and urban centres are, then conditioning on time-varying climate, emissions, and human activity---mirrors the physical intuition that slow-changing geographic structure constrains the faster dynamics that unfold on top of it. Static features can also be concatenated as additional channels alongside temporal variables in a single-stage training setup.

\subsection*{Temporal coverage and known limitations}
Temporal coverage is intentionally heterogeneous across domains and reflects the availability of the underlying source products rather than a design flaw. Climate variables extend back to 1940, land use to 1900, and energy infrastructure to 1900, but ocean biogeochemistry begins only in 2010, air quality in 2003, and vegetation in 2000. Within a given year, some variables will be present while others will not. Users building multi-domain training tensors should account for this by either restricting to the intersection of available years across selected variables, or by implementing missing-variable masking strategies that allow the model to learn from partial observations.

The annual resolution makes WorldTensor well suited to applications driven by interannual variability, long-term trends, and cross-domain coupling: pretraining geospatial foundation models, learning joint human--environment representations, transfer learning across socioeconomic and environmental targets, feature generation for climate-impact and risk assessment, and analyses of policy-relevant decadal change. It is correspondingly less suited to applications requiring sub-annual fidelity---operational or sub-daily weather forecasting, seasonal-cycle and monthly-anomaly analysis, and event-level detection of short-lived extremes---for which the original sub-annual products in Table~\ref{tab:sources} should be used directly. As shown in our temporal validation, signals that unfold and recover within a single year (e.g.\ the 2020 COVID-19 emissions dip) are attenuated in the annual aggregates.

The $0.25^\circ$ grid represents a compromise between spatial detail and global consistency. For variables derived from high-resolution remote sensing (e.g.\ settlement layers at 10--30\,m), the regridding to ${\sim}28$\,km cells averages out fine-grained urban structure. Conversely, for variables derived from sparse point observations (e.g.\ power plants, conflict events), the $0.25^\circ$ grid may introduce apparent spatial precision beyond the effective resolution of the underlying data. Users should interpret grid-cell values in light of the native resolution documented for each source.

\section*{Code availability}

All code used to acquire, harmonise, and quality-check the WorldTensor dataset is publicly available on GitHub at \url{https://github.com/crp94/worldtensor-pipeline} and archived on Zenodo (\href{https://doi.org/10.5281/zenodo.19184043}{doi:10.5281/zenodo.19184043}). The repository includes source-specific download and preprocessing scripts, the spatial and temporal harmonisation pipeline, a post-hoc structural compliance tool, and PyTorch dataset interfaces for model training. The repository is released under the MIT License. Python~3.10+ with \texttt{xarray}, \texttt{netCDF4}, \texttt{rasterio}, and standard scientific computing libraries is required; PyTorch examples additionally require \texttt{torch}, \texttt{torchgeo}, and \texttt{scikit-learn} (installable via the \texttt{ml} optional dependency group).

\section*{Data availability}

The WorldTensor dataset is available on Zenodo at \href{https://doi.org/10.5281/zenodo.19047618}{doi:10.5281/zenodo.19047618} \cite{worldtensor_data} under a Creative Commons Attribution 4.0 International (CC~BY~4.0) license. The release comprises 52{,}823 NetCDF files organised into 14 thematic domains plus a static context domain, totalling approximately 46~GB. All files follow the spatial and metadata conventions described in this paper. Individual domains or variable families can be downloaded independently. Source datasets used to construct WorldTensor are publicly available from the repositories cited in Table~\ref{tab:sources}; users should consult the original licenses for any use beyond the scope of this release.

\bibliography{bibliography}

\section*{Acknowledgements}

Authors acknowledge support from the European Research Council, ERC grant agreement number 101044703 (EUNICE) CUP D87G22000340006.

\section*{Contributions}

C.R.-P.\ conceived the project, designed the data model and harmonisation framework, implemented all processing pipelines, performed the validation analyses, and wrote the manuscript. M.T.\ supervised the project, provided scientific guidance, and reviewed the manuscript.

\section*{Ethics declarations}

\subsection*{Competing interests}
The authors declare no competing interests.




\end{document}